\documentclass[letterpaper, 10 pt, journal, twoside]{IEEEtran}  % Comment this line out if you need a4paper

\IEEEoverridecommandlockouts                              % This command is only needed if 
\usepackage{multirow}

\usepackage[noend]{algpseudocode}
\usepackage{algorithm}
\usepackage{amssymb}
\usepackage{bbm}
\usepackage{dsfont}
\usepackage{makecell}
\usepackage[left=1.69cm,right=1.69cm,top=2.12cm,bottom=1.52cm]{geometry}

\usepackage{xspace}
\usepackage{xcolor}
\definecolor{softgreen}{RGB}{34,139,34} % A visually appealing forest green
\usepackage[colorlinks=true, linkcolor=softgreen, citecolor=softgreen, urlcolor=softgreen]{hyperref}
\usepackage{array}

\newcommand{\X}{\ensuremath{\mathcal{X}}\xspace}
\newcommand{\T}{\ensuremath{\mathcal{T}}\xspace}
\newcommand{\Xg}{\X_G} 
\newcommand{\xs}{\ensuremath{\x_\start}} 
\newcommand{\U}{\ensuremath{\mathcal{U}}\xspace}
\newcommand{\uu}{u}
\newcommand{\Xfree}{\ensuremath{\X_{\text{free}}}\xspace}
\newcommand{\Xobs}{\ensuremath{\X_{\text{obs}}}\xspace}
\newcommand{\K}{\ensuremath{I}\xspace}
\newcommand{\Kspace}{\K-space\xspace}
\newcommand{\key}{\ensuremath{\mathbb{K}}\xspace}
\newcommand{\skey}{\ensuremath{\mathds{k}}\xspace}
\newcommand{\keys}{\ensuremath{\key_\text{start}}\xspace}
\newcommand{\keyg}{\ensuremath{\key_\text{goal}}\xspace}
\newcommand{\Kobs}{\ensuremath{\K_{\text{obs}}}\xspace}
\newcommand{\Kfree}{\ensuremath{\K_{\text{free}}}\xspace}
\newcommand{\x}{x}

\renewcommand{\xi}{\x_{I}}
\newcommand{\prm}{\textsc{prm}\xspace}
\newcommand{\lprm}{\textsc{lazy-prm}\xspace}

\newcommand{\start}{\text{start}}

\newcommand{\goal}{\text{goal}}

\newcommand{\im}{\ensuremath{Im}\xspace}
\newcommand{\IM}{\ensuremath{\mathcal{IM}}\xspace}

\newcommand{\C}{\X}

\newcommand{\Cf}{\Xfree}
\newcommand{\Co}{\Xobs}

\usepackage[textsize=small]{todonotes}
\usepackage{graphics} % for pdf, bitmapped graphics files
\usepackage{listings}
\usepackage{comment}
\usepackage{epsfig} % for postscript graphics files
\usepackage{amsmath} % assumes amsmath package installed
\usepackage{cite}
\usepackage{subcaption}
\usepackage{xcolor}
 \usepackage{url}
\definecolor{gg}{RGB}{0, 155, 85} 
\definecolor{json-key}{rgb}{0.13,0.55,0.13}
\definecolor{json-value}{rgb}{0.25,0.25,0.25}
\definecolor{json-string}{rgb}{0.9,0.3,0.3}
\usepackage{caption}
\lstdefinelanguage{json}{
  basicstyle=\ttfamily,
    commentstyle=\color{gray},
    stringstyle=\color{orange},
    numbers=left,
    numberstyle=,
    numbersep=5pt,
    breaklines=true,
    frame=lines,
    backgroundcolor=\color{white!10},
    captionpos=b
}

\newcommand\centerImage[2][]%
  {\raisebox
    {-\dimexpr0.5\height}%
    [\dimexpr0.5\height+1mm]%
    [\dimexpr0.5\height+1mm]%
    {\includegraphics[#1]{#2}}%
  }

\title{Image-Based Roadmaps for Vision-Only Planning and Control of Robotic Manipulators}

\author{{Sreejani Chatterjee, Abhinav Gandhi, Berk Calli, Constantinos Chamzas}
\vspace*{-1.65em}
\thanks{Manuscript received: February, 11, 2025; Revised May, 10, 2025; Accepted June, 13, 2025.}
\thanks{This paper was recommended for publication by Editor Dr. Aniket Bera upon evaluation of the Associate Editor and Reviewers' comments.
This work was supported by the National Science Foundation under Award No. 2341532}
\thanks{\footnotesize{All authors are
 with the Robotics Engineering Department of Worcester Polytechnic Institute,
 Worcester, MA 01609, USA.} {\tt\footnotesize schatterjee\@wpi.edu/sreejani.c@gmail.com}}%
\thanks{\footnotesize{Digital Object Identifier (DOI): see top of this page.}}
}

\begin{document}
\maketitle
\markboth{IEEE Robotics and Automation Letters. Preprint Version. Accepted June, 2025}
{Chatterjee \MakeLowercase{\textit{et al.}}: Vision-only motion planning and control} 
%%%%%%%%%%%%%%%%%%%%%%%%%%%%%%%%%%%%%%%%%%%%%%%%%%%%%%%%%%%%%%%%%%%%%%%%%%%%%%%%
\begin{abstract}

This work presents a motion planning framework for robotic manipulators that computes collision-free paths directly in image space. The generated paths can then be tracked using vision-based control, eliminating the need for an explicit robot model or proprioceptive sensing.
At the core of our approach is the construction of a roadmap entirely in image space. To achieve this, we explicitly define sampling, nearest-neighbor selection, and collision checking based on visual features rather than geometric models. We first collect a set of image space samples by moving the robot within its workspace, capturing keypoints along its body at different configurations. These samples serve as nodes in the roadmap, which we construct using either learned or predefined distance metrics.
At runtime, the roadmap generates collision-free paths directly in image space, removing the need for a robot model or joint encoders. We validate our approach through an experimental study in which a robotic arm follows planned paths using an adaptive vision-based control scheme to avoid obstacles. The results show that paths generated with the learned-distance roadmap achieved 100\% success in control convergence, whereas the predefined image space distance roadmap enabled faster transient responses but had a lower success rate in convergence.
 
\end{abstract}
\vspace{-0.5em}

\vspace{-1.25em}

\section{Introduction}
\vspace{-0.35em}
\IEEEPARstart{V}{ision-based} control techniques~\cite{hutchinson1996tutorial, hashimoto2003review}, offer significant advantages for robotic manipulators in unstructured and cluttered environments by enabling closed-loop control using task-relevant visual information. These techniques also enhance robustness against model inaccuracies, beneficial for robots with complex or variable dynamics \cite{Calli2016, cuevas2018hybrid, ardon2018reaching}. This strategy is especially useful for robots that are difficult to model accurately, e.g. soft robots \cite{Lai2020, luo2018orisnake}, under-actuated robots \cite{liu2020survey, gandhi2023shape}, 3D printed robots \cite{chavdarov2019design, onal2014origami}, and robots with inexpensive hardware \cite{adzeman2020kinematic}. Furthermore, model-free visual servoing, which learns robot-feature motion models during control, reduces reliance on a priori knowledge of the robot model \cite{wang2018adaptive, navarro2017fourier}.

The goal of our research is to push the boundaries of purely vision-based control and motion planning for robotic manipulators, by decreasing reliance on explicit robot modeling or proprioceptive sensing. In doing so, we strive to use natural visual features along the robot's body in image space, without attaching any external markers. We use multiple features along the robot's body to uniquely distinguish full-body configurations that have the same end effector position, a characteristic common in redundant robots such as continuum manipulators. While the works in \cite{gandhi2022skeleton,chatterjee2023keypoints, chatterjee2024utilizing} provided algorithms to track natural keypoints and use them to achieve vision-based model-free control with decent transient responses, these algorithms are only designed to run in obstacle-free space and do not provide motion planning capability to avoid obstacles in the scene. 

\begin{figure}[t]
        \centering
    \includegraphics[width=0.65\columnwidth]{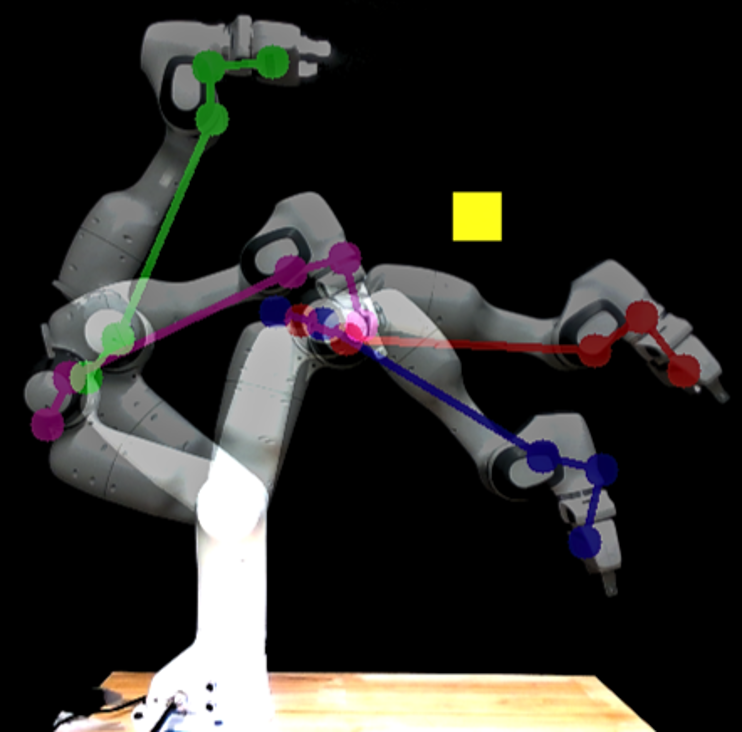}
    \caption{
    The robot follows a collision-free path using visual keypoints (colored) planned with the proposed method. 
    The set of keypoints are tracked in order:
    green, magenta, blue, and red. The robot avoids the yellow obstacle while moving between the specified start and goal.  }
    \label{intro-image}
\end{figure}

In this work, we tackle the problem of collision-free path planning for purely vision-based robot control: we develop a planning and control scheme that does not rely on a robot model or on-board sensors (e.g. joint encoders) in runtime. As such, this strategy is especially useful for (but not limited to) soft/underactuated/inexpensive robots that are hard to model and may not have reliable proprioceptive sensing.

Toward this goal, we propose a novel motion planning formulation that operates only with visual information. We propose a vision-only motion planning methodology using roadmaps \cite{kavraki1996probabilistic}.  We investigated two ways of generating roadmaps using natural features on a robot: 1) directly utilizing the Euclidean distance between keypoints in image space as a distance metric, 2) estimating the joint displacements from image features and utilizing it as a distance metric. For the latter, we first used an automated data collection pipeline to annotate natural keypoints' placement in image space along the robot's body as the robot moves across various configurations.

This dataset was used to develop a simple neural network that approximates joint displacements based on keypoint locations in image space. These distance metrics were integrated into the roadmap construction. Once the roadmaps are constructed, polygon-based collision checking and A* \cite{candra2020dijkstra} search are employed to ensure collision-free paths. These two approaches have different implications for vision-based robot control. In a nutshell, we observed that utilizing estimated joint distances in the roadmap results in smoother and more accurate tracking of the generated path, allowing the robot to stay in its defined workspace and avoid obstacles. In contrast, roadmaps based on Euclidean distances in image space can offer faster transient responses, albeit with potentially less reliable tracking. in environments with tightly spaced or irregularly shaped obstacles, these image space roadmaps frequently fail to generate feasible paths due to their poor alignment with the underlying joint-space feasibility. Our experiments demonstrate these aspects both in the presence and absence of obstacles.
 
\section{Related Work}
\vspace{-0.5em}
Motion planning is a core problem in robotics that has been extensively studied over the decades. It is generally categorized into three main types: optimization-based \cite{schulman2014motion, zhao2024survey}, sampling-based \cite{orthey2024review, kingston2018sampling}, and search-based planners \cite{likhachev2003ara, cohen2010search}. Recently, Jacobian-based motion planning \cite{park2020trajectory} has also gained traction for obstacle avoidance tasks. While all methods have found widespread success in different applications, all of these rely on having an explicit geometric model of the robot to design feasible paths. In this work we extend the principles of sampling-based planning of the Probabilistic Roadmap (\prm) approach \cite{kavraki1996probabilistic}, to provide a method for motion planning where a robot model is not available. 

Model-free planning, especially without prior knowledge of the robot's geometry, presents unique challenges. Reinforcement learning (RL) has been explored extensively in the recent decades to address such challenges. For instance, \cite{liu2021model} proposed an RL-based framework for generating jerk-free, smooth trajectories. While effective, this method depends on carefully crafted reward functions and extensive datasets generated in simulation, with no real-world validation. Similarly, \cite{zhou2021robotic} introduced a hybrid approach combining RRT*-based trajectories with PPO reinforcement learning for policy refinement. However, this method heavily relies on model-based elements like precomputed trajectories and supervised learning for initial policy design, with experiments confined to simulated environments. In contrast, our approach eliminates reliance on explicit or precomputed models and trajectories by leveraging visual keypoints to construct roadmaps directly in image space, requiring only a small dataset collected from a real robot. This makes our method more adaptable to scenarios without precise geometric models.

Among related works, the approaches proposed in \cite{ichter2019robot} and \cite{556169} are the most closely aligned with ours. In \cite{ichter2019robot}, the authors plan trajectories by learning a low-dimensional latent space using three neural networks: an autoencoder, a latent dynamics model, and a collision checking network. This latent space allows planning without directly operating in the high-dimensional input space (e.g., raw images or complex joint configurations). While effective, their method requires extensive simulation-based training, which introduces a significant sim-to-real gap. Moreover, the system is validated only in simulation, and the reliance on multiple learned modules increases complexity, making real-world deployment more challenging. In contrast, our method operates directly in image space using visually tracked keypoints, avoiding the need for encoding, decoding, or learned dynamics. It uses real robot data exclusively, both for training and evaluation, and does not rely on simulation. This results in a lightweight, practical framework that is easier to implement and deploy in real-world scenarios. In \cite{556169}, the authors learn the topology of a perceptual control manifold (PCM) using a topology-preserving neural network (TRN) constructed over both robot joint configurations and corresponding image features. This approach explicitly relies on access to \textbf{joint space} information and was designed specifically for a pneumatically driven robotic arm, limiting its generalizability. In contrast, our method requires only a dataset of image features and does not depend on any explicit robot model or configuration information. As a result, our approach is more general and can be easily modified to both rigid and soft robotic manipulators.

Planning a path for image based visual servoing without a-priori knowledge of the robot's model is rarely delved into in the literature. For instance, \cite{mezouar2000path} modeled a path planner for visual servoing to bridge gaps between initial and target positions which are much further apart in configuration space, without addressing collision avoidance. In \cite{lee2011obstacle} authors achieve obstacle avoidance in pose based visual servoing and hence still needed explicit robot model instead of only visual feedback.
\vspace{-0.75em}
\section{Preliminaries and Problem Statement}
\vspace{-0.35em}
In this section we describe the motion planning problem, a brief review of sampling-based methods focusing on probabilistic roadmaps, and finally we introduce the problem statement that we are trying to solve.  

\begin{figure}[t]
      \centering
      \includegraphics[width=0.5\textwidth]{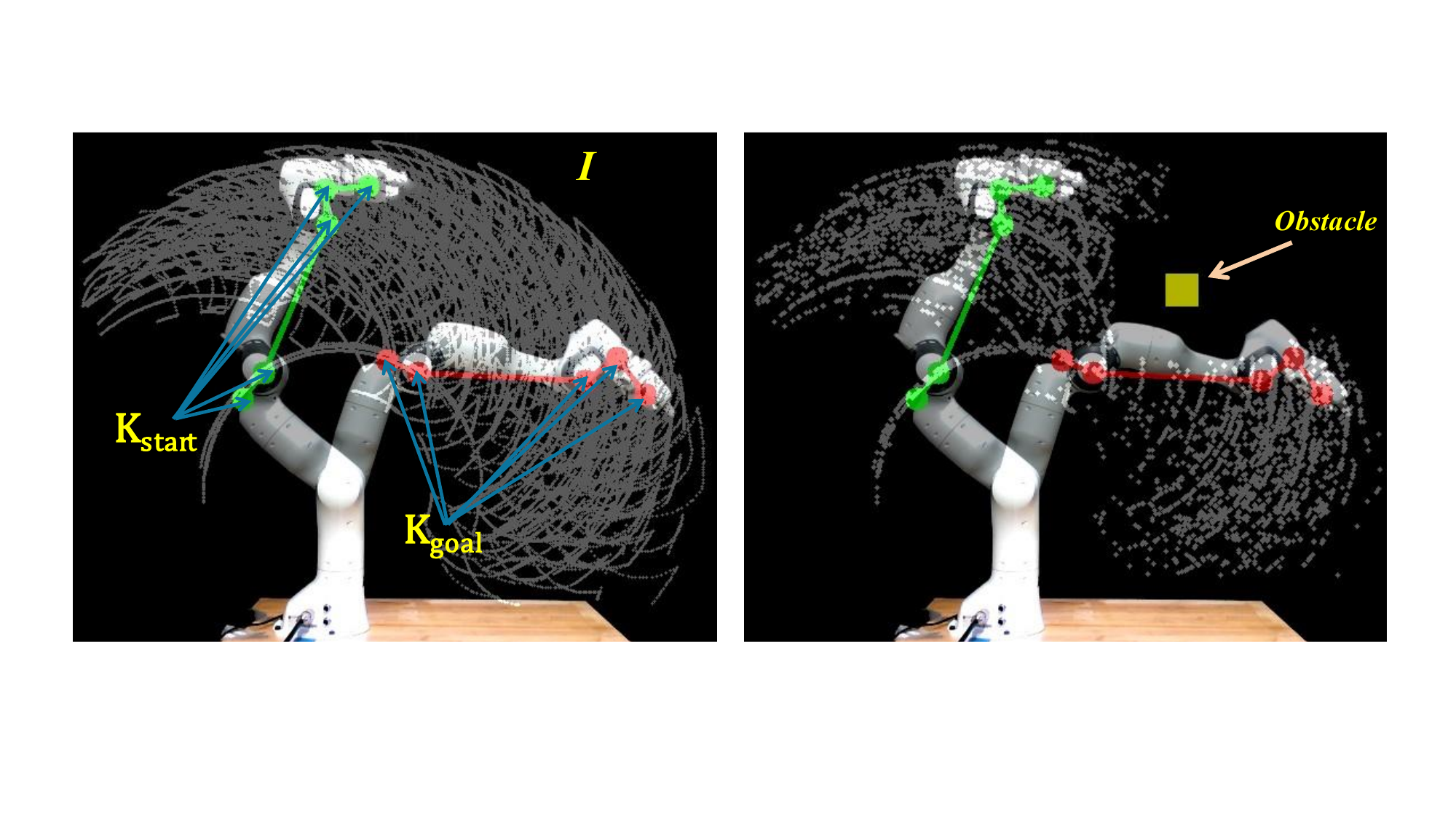}
      \vspace{-4em}
      \caption{The images depict a vision-based motion planning problem. The left image shows the discretized representation of \K  highlighted in gray, where each configuration is sampled as an image frame, resulting in a finite set of \key. The gray region in the right image highlights the same discretized representation of \Kfree, avoiding the obstacle in yellow. \Kfree ensures a collision-free path from the start configuration \keys to the goal configuration \keyg}
      \label{config_space}
\vspace{-1.5em}
\end{figure}

\vspace{-1.5em}
\subsection{Model-based Motion Planning}
\label{ssec:prob-state}

Let $\x \in \X$ denote the state and state-space, and $ \uu \in \U$ the control and control-space of a robot system \cite{orthey2024review}.
The system dynamics equations $f$ of the robot system can be written as 
\begin{equation} \label{eq:motion}
\footnotesize
\x(\T) = \x(0)+\int_0^{\T} f(x(t),u(t))dt \end{equation}
If the system dynamics impose constraints on the allowable paths the system is called a non-holonomic system.  Now, Let $\Co \subset \C $ denote the invalid state space, which is the set of states that violates the robots kinematic constraints or collides with the workspace $\mathcal{W} \subseteq \mathbb{R}^3$. e.g.,
\begin{equation}
\footnotesize
\Co = \{ \x \in \C \mid R(\x) \bigcap \mathcal{O} \neq \emptyset \}
\end{equation}

where $R(\x) \subseteq \mathbb{R}^3$ is the set of all points occupied by the robot at state $\x$, and $\mathcal{O} \subset \mathcal{W}$ denotes the set of obstacles within the workspace\footnote{This equation only encodes collisions; joint or kinematic constraints can be incorporated in a similar manner.}. Let $\Cf = \C \setminus \Co$ denote the collision-free state space. Also, let $x_\start \in \Cf$ and $\Xg \subseteq \Cf$ represent the start state and goal region, respectively.

\textit{Motion Planning Problem:}
Given the motion planning tuple $(\X, R, \mathcal{W}, \U, f,  x_\start, \Xg)$ find a time $\T$ and a set of controls $u: [0,\T] \rightarrow \U$ such that the motion described by \autoref{eq:motion} satisfies $ x(0) = \xs $, $ \x(\T) \in \Xg$ and $x(t) \in \Xfree$.   

\label{ssec:prob-state-vis}

The problem we are considering in this setting is departing from the above motion planning problem as the geometric model $R(x)$, the dynamics $f$ and the workspace $\mathcal{W}$ is not directly available. Instead the only information available is the space of admissible controls \U and an image $\im \in \IM$.

We will describe the new problem statement by explicitly defining equivalent concepts of obstacles, robot states, and state dynamics directly in the image space.
We assume that we are given a point tracking function that maps a given image to $N$ fixed pixel points on the robot's body.  
This vector of pixel points is denoted as an image state $\key \in \K $.
Where $\K \subset \mathbb{R}^{N \times 2}$ denotes the space of all image states.
In \autoref{intro-image} we can see $4$ different robot configurations in image space. Each configuration is represented by a set of $5$ same-colored circles. Each of these circles on the robot's body in the image is a pixel point and is denoted as a keypoint or $\skey$. The kinematic chain or shape formed by the vector of $5$ circles represents one image state \key.

Given a set of image obstacles $\mathcal{IO} \subset \im$ we define: 
\vspace{-0.5em}
\begin{equation}
\footnotesize
\Kobs =\{\key \in \K | RI(\key) \bigcap \mathcal{IO} \neq \emptyset \}
\vspace{-0.5em}
\end{equation}
where $RI(\key) \subseteq \im $ is the set of pixels that the robot occupies at image state \key . Similarly to before, we define $\Kfree  = \K \setminus \Kobs$, $\key_\start$ and $\K_g$. 

We define the unknown system dynamics function as:  
\vspace{-0.5em}
\begin{equation} \label{eq:motion_image}
\footnotesize
\key(\T) = \key(0)+\int_0^{\T} g(\key(t),u(t))dt
\vspace{-0.5em}
\end{equation}
Here we note that even if the underlying function f is fully-integrable (holonomic-system) if $dim(\K)> dim(\X)$ the equivalent system dynamics equations $g$ will be non-holonomic as there will be paths in the higher dimensional image space \K that cannot be followed by the robot. In our approach we consider this issue, and propose a way to produce paths that approximately satisfy these constraints. In the first image of \autoref{config_space}, the grey region illustrates the discretized representation of \K, as each configuration is captured as an image frame, resulting in a finite set of \key sampled at a fixed frame rate. The second image of \autoref{config_space} highlights similar discretized representation of \Kfree.

 Given the above definitions, let us define the vision-only motion planning problem. 
\textit{Vision-Only Motion Planning Problem:}
Given the tuple $(\K, RI, \U, \mathcal{IO}, \im, \key_{\start}, \K_g)$ find a time $\T$ and a set of controls $u: [0,\T] \rightarrow \U$ such that the motion described by $g$ satisfies $\key(0) = \key_{\start}$, $ \key(\T) \in \K_g$ and $\key(t) \in \Kfree$.  

\section{Methodology}
\vspace{-0.25em}
\label{ssec:methodology}
Since the system dynamics equation is unknown, we cannot directly plan in the control space \U. Instead, we will plan a path $\key_0, \key_1 \ldots, \key_n$ directly in the \Kspace and then control the robot to follow the path with a vision-only controller\cite{gandhi2022skeleton}.

To compute the path, which is the main contribution of this work, we propose to use probabilistic road-map planner (\prm) by adapting its subroutines to operate directly in \K-space. Specifically, we opted to adapt the \lprm~\cite{bohlin2000path} planner due to its compatibility with our particular requirements. However, any sampling-based planner which relies on the same subroutines could be used. 

\vspace{-1.0mm}
\begin{algorithm}[H]
\footnotesize
   \caption{Build-Lazy-PRM} 
   \label{alg:build-lazy-prm}
    \begin{algorithmic}[1] 
     \Procedure{Build-Lazy-PRM}{N, k}  
        \State {$G$} $\gets$ INIT()   
        \While{$G$.size()$\leq$ N}%$i = 1 ,\ldots, N$} 
          %\State \Comment{generate collision free sample}
           \State $\key_{new} \gets $ {\color{red}{\textsc{sample}}}($\K$) \label{sample} 
           \State $G$.addNode($\key_{new}$) \label{roadmap}
        \EndWhile
       \For{each $\key \in  G$.nodes()} 
           \State $\mathcal{N}(\key) \gets $ {\color{red}{\textsc{K-nearest}}}($\key$, {$G$}) \label{dist} 
           \For{each $\key_{near} \in \mathcal{N}(\key_{new})$} 
            \State $e \gets$ ($\key, \key_{near})$
            \If{$e \notin G$.edges()} \label{collision}
                \State  $G$.addEdge($e$) 
            \EndIf 
          \EndFor 
        \EndFor 
    	\State \Return $G$ 
    \EndProcedure
    \end{algorithmic}
    \end{algorithm}
\vspace{-1.5em}
\begin{algorithm}[H]
\footnotesize
   \caption{Query-Lazy-PRM} 
   \label{alg:query-lazy-prm}
    \begin{algorithmic}[1] 
     \Procedure{Lazy-Query-PRM}{\keys, \keyg, $G$}   
       \State G.add(\keys) \Comment{Add start, and goal to the Graph}
       \State G.add(\keyg)
       \State Edges, $\gets$ \textsc{Search-Graph} $G$(\keyg, \keys)
       \For{each $e \in  G$.edges()} 
           \If{{\color{red}{\textsc{coll\_free}}}($e$)} \label{collision}
           \For{each $\key_{near} \in \mathcal{N}(x_{new})$} 
            \State $e \gets$ ($\key, \key_{near})$
            \If{{\color{red}{\textsc{coll\_free}}}($e$) and $e \notin G$.edges()} \label{collision}
                \State  $G$.addEdge($e$) 
            \EndIf 
          \EndFor 
    	\State \Return $Path$ 
        \EndIf 
     \EndFor 
    \EndProcedure
    \end{algorithmic}
    \end{algorithm}
    \vspace{-0.75em}

Similar to \prm, \lprm operates in two phases: a building phase \autoref{alg:build-lazy-prm}, where the roadmap is constructed without collision checking, and a query phase \autoref{alg:query-lazy-prm}, where a path is searched and only the selected edges are checked for collisions. If any are invalid, the roadmap is updated and the search is repeated. This separation allows the roadmap to be built offline, before robot deployment, without considering obstacles, enabling reuse across different environments and physical setups.

We use the probabilistic roadmap method for its efficiency in multi-query scenarios, where a single precomputed graph supports multiple start–goal pairs. The lazy variant is particularly suitable as the environment remains mostly unchanged between planning queries. Its ability to handle high-dimensional spaces also makes it well-suited for our \K-space representation based on visual keypoints. We first describe how \lprm works, followed by the modifications needed to adapt it to \K-space.

The building-phase (\autoref{alg:build-lazy-prm}) works with the following procedure. First, in line \ref{sample}, a sample is generated and added in graph $G$. Then the k nearest neighbors are found by using a distance defined in \K (line \ref{dist}) and are connected with edges. This continues until $N$ nodes are in the graph $G$. 

During the query-phase (\autoref{alg:query-lazy-prm}) a new motion planning problem is solved. Given a $\key_\start$ and $\key_\goal $ they are added in the  graph, and connected with their nearest-neighbors. Then a graph search algorithm e.g., A* is used to find a path. If edges of the path are in-collision they are updated accordingly in the roadmap, and the process repeats until a collision free-path is found. 
The three operations \textsc{sample},  \textsc{k-nearest},  \textsc{coll\_free},
for \autoref{alg:build-lazy-prm} in lines \ref{sample}, \ref{dist} and \autoref{alg:query-lazy-prm}  typically require a model for the robot.

\begin{figure*}[ht!]
      \centering
      \includegraphics[width=0.7\textwidth]{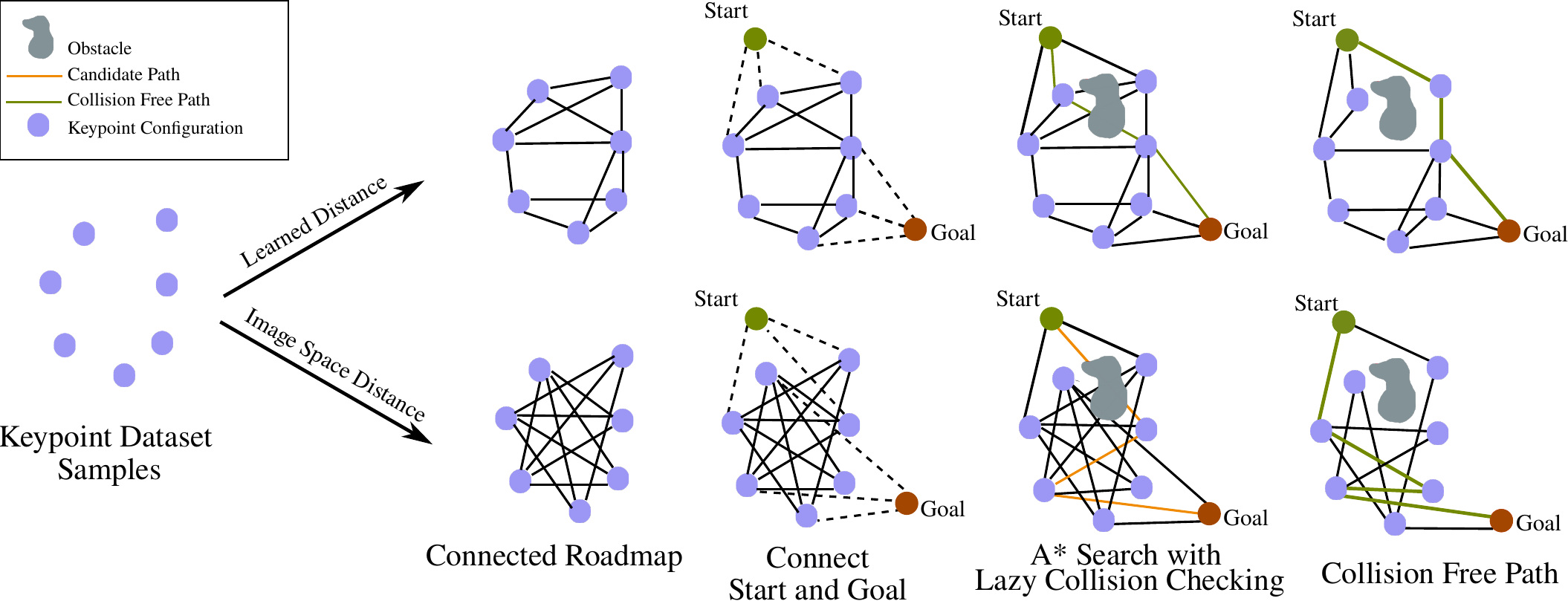}
          \caption{Overview of the roadmap creation process for motion planning. In this image the purple circles are nodes in image space represented by image state in each frame. Using a distance metric, we connect the nodes to create a roadmap. As observed two different metrics produce different edges on the graph. After an A* search for path finding between a pair of start (\keys) and goal (\keyg) configurations and collision check if required, paths are found for both roadmaps. The green lines denote the final path after collision check. As observed, the \textbf{learned} distance roadmap has a clearly more optimized path than the \textbf{image space} distance roadmap}
      \label{graph}
      \vspace{-1.4em}
\end{figure*}

\textsc{sample}: This function typically samples the configuration space uniformly. However, in the vision-only setting, we can't directly sample in \Kspace as we don't have the model of the robot. Since $dim(\K) > dim(\X)$ the keypoint vectors (image state \key) that correspond to actual configurations of the robot, will lie in a lower-dimensional (equal to $dim(\X)$) manifold in \Kspace. Thus 
randomly sampling \Kspace will have $0$ probability of sampling a valid configuration that lies on the valid manifold \cite{kingston2018sampling}. We describe how to address this issues in \autoref{ssec:dataset}, by collecting and storing valid samples directly from the real robot.  

\textsc{k-nearest}: This function usually relies on a distance defined in \C  and finds the nearest configurations that can be connected. However, since our representation in \Kspace is now a non-holonomic system, defining this distance is very challenging, as a straight line path in \Kspace defined by a simple Euclidean distance, might not be accurately followable by a controller. To mitigate this, we describe a learning-based approach to estimate the unknown joint-distance, in \autoref{ssec:dist_metric}.

\textsc{coll\_free}: This function checks if there is a collision for an edge in \C. Again this typically requires the model $R(x)$ for the robot. In the vision only case, we propose simple yet successful method to do collision checking with an $RI(x)$ directly in \Kspace \autoref{ssec:line-check}.     

In the next section we describe our proposed method for each of the aforementioned subroutines. \autoref{graph} visually describes all the steps of the visual motion planning framework.

\subsection{\textsc{sample}: Executed Trajectories as Proxy Samples} 
\vspace{-0.35em}
\label{ssec:dataset}
In the absence of a robot model, we represent each image state $\key \in \K$ by identifying and annotating keypoints (\skey)
 on the robotic arm using an automated data collection pipeline described in \cite{chatterjee2023keypoints, chatterjee2024utilizing}. Each \ensuremath{\key_n} is composed of a set of $N$ keypoints \skey, where each \skey represents a pixel point of a specific location on the robot's body in image space. For instance, if \( N = 5 \), a keypoint vector or image state \ensuremath{\key_n} is represented by a vector of {\ensuremath{\skey_1}, \ensuremath{\skey_2}, \ensuremath{\skey_3}, \ensuremath{\skey_4}, \ensuremath{\skey_5}}, where each \skey is a pixel point in the $n\textsuperscript{th}$ image frame. An example vector of such keypoints is shown in \autoref{config_space} as \keys or \keyg

 To systematically explore the robot's visible workspace in the image space and ensure comprehensive coverage, we compute velocities for each joint \( j \) using:

\begin{equation}
\footnotesize
\label{comp_vel}
v_j = \min\left(\frac{M_j}{\ensuremath{res} \cdot dt}, v_{\text{max}}\right),
\end{equation}

where, \( M_j \) is the motion range or difference of limits of joint \( j \), \ensuremath{res}, is the number of discrete  steps used to traverse \( M_j \), \( dt \) is the duration allocated to complete each step, and \( v_{\text{max}} \) is the maximum allowable velocity for joint \( j \).
By dividing the motion range of each joint into uniform increments based on \ensuremath{res}, we ensure that the robot systematically explores all possible configurations within its workspace. The resulting dataset of image states \key is thus evenly distributed across the image space \K, enabling robust coverage and accurate representation of the robot's motion capabilities. 

Each configuration \ensuremath{\key_n} is computed using the following transformation:
\vspace*{-1.5mm}
\begin{equation}
\footnotesize
\label{2d_eq}
\ensuremath{\key_n} = K \cdot T_{cw} \cdot \ensuremath{x_n}
\vspace{-0.75em}
\end{equation}
where $K$ is the camera intrinsic matrix, $T_{cw}$ is the camera extrinsics matrix derived from calibration processes described in \cite{Lee2020, Zhang2000},  and \ensuremath{x_n} is the 3D robot configuration in workspace. 
This transformation projects the 3D configuration \ensuremath{x_n}, into their corresponding 2D (pixel) projection in the image described in \cite{3dRecon}, resulting in a set of \skey for each image state \ensuremath{\key_n} The process captures the robot's motion across its visible workspace and creates a comprehensive dataset of \key, representing close to all feasible configurations in image space. Dataset of \ensuremath{\key} can also be collected by following the process describe in \cite{chatterjee2024utilizing}. 

The \textbf{Keypoint Dataset Samples} section of \autoref{graph}, represents this part of the workflow. The grey area in the left image of \autoref{config_space} illustrates how \key are distributed within image space with each frame consisting of a set of \skey or keypoints. This collection process yields a dataset of \key that can be used to enable efficient roadmap construction for vision-based motion planning.

Since this data collection process relies only on observable image features, it can be easily applied to other robotic systems, including soft and rigid manipulators, enabling broad applicability of the proposed framework.
\vspace{-1.25em}
\subsection{\textsc{k-nearest}: Learned and Image Space Metrics}
\vspace{-0.25em}
\label{ssec:dist_metric}
To search for the K-nearest neighbors of each image state sample we employ the following two distance metrics:

\begin{itemize}
\item \textit{Learned distance},  where the distance is learned by a neural network, trained to predict joint displacements between two image states. The input to the network is a pair of image states $\key_1$ and $\key_2$ and the output is an estimated joint displacement required to transition between them:
\begin{equation}
\footnotesize
\label{cust-eq}
\text{dist}_{learned}() \leftarrow \text{NN}(\key_1, \key_2)
\end{equation}
\item \textit{Distance in Image space}, where we simply calculate the Euclidean distance between $\key_1$ and $\key_2$ in image space:
\begin{equation}
\footnotesize
\label{euc-eq}
\text{dist}_{image}() \leftarrow || \key_1 - \key_2 ||_2
\end{equation}
\end{itemize}
Each metric influences the graph's structure by determining the nearest neighbors and defining the edges in the roadmap. The learned distance prioritizes image states with minimal estimated joint displacement, while the image space distance favors image states that are closer in the image. These differences impact the connectivity of the roadmap, as illustrated in the \textbf{Connected Roadmap} and \textbf{Connect Start and Goal} sections of \autoref{graph}.

\subsubsection{Dataset generation for network model}
\label{ssec:approx-joint}
While collecting the dataset of image states \key in \autoref{ssec:dataset}, we recorded the velocity (\autoref{comp_vel}) applied to transition between consecutive frames. Using this data, we created a new dataset that includes pairs of consecutive image states (\ensuremath{\key_1}, \ensuremath{\key_2}) and their estimated joint displacements. This is calculated by multiplying the recorded velocity with the frame rate at which \key was captured, as described in \autoref{2d_eq}.

 Only pairs of \key captured in consecutive image frames are included in this data generation process.  However, constructing the graph $G$, requires computing distances between arbitrary \key pairs in \K-space. To achieve this, a neural network is trained to predict joint displacements between different image-state pairs \key.
 
To enhance diversity, the dataset is augmented by combining frame sequences where the first frame's image state (\(\key_{\text{start}}\)) and the last frame's image state (\(\key_{\text{end}}\)) act as boundaries, with total joint displacements calculated as sum of displacements across intermediate frames. This approach ensures diverse transitions, enabling the neural network to accurately estimate joint displacements for any image state pair.

\subsubsection{Neural Network for Learned Distance Metric}
\label{ssec:reg-model}
To derive \autoref{cust-eq}, we design a simple neural network using the aforementioned dataset to learn a distance more similar to the \textbf{joint space} distance. The model takes concatenated arrays of the starting image state (\(\key_{\text{start}}\)) and the subsequent image state (\(\key_{\text{next}}\)) as input and predicts the estimated joint displacement. We train the network using the Adam optimizer (learning rate: 0.005), a batch size of 32, and 400 epochs, minimizing mean squared error (MSE) loss. Performance is evaluated using RMSE, MAE, and \( R^2 \) metrics.

\vspace{-1.75em}
\subsection{\textsc{coll\_free}:Image-Based Collision Checking}
\label{ssec:line-check} 
\vspace{-0.35em}

To enable collision checking in our model-free system, we employ an image-based polygon collision-checking framework. Obstacles are defined directly in the image space by drawing contours around visible objects and expanding them with a safety margin to account for the robot’s volume and controller uncertainty. Each candidate path is checked by testing for intersection between the expanded obstacle region and polygons formed by corresponding keypoint pairs across two image states. This method effectively handles obstacles of arbitrary shape and size and does not require access to the robot’s model or workspace geometry
The collision checking process and \textbf{A*} search is shown in  \textbf{A* Search with Lazy Collision Checking} of \autoref{graph}.  

\begin{figure}[t]
      \centering
      \includegraphics[width=0.475\textwidth]{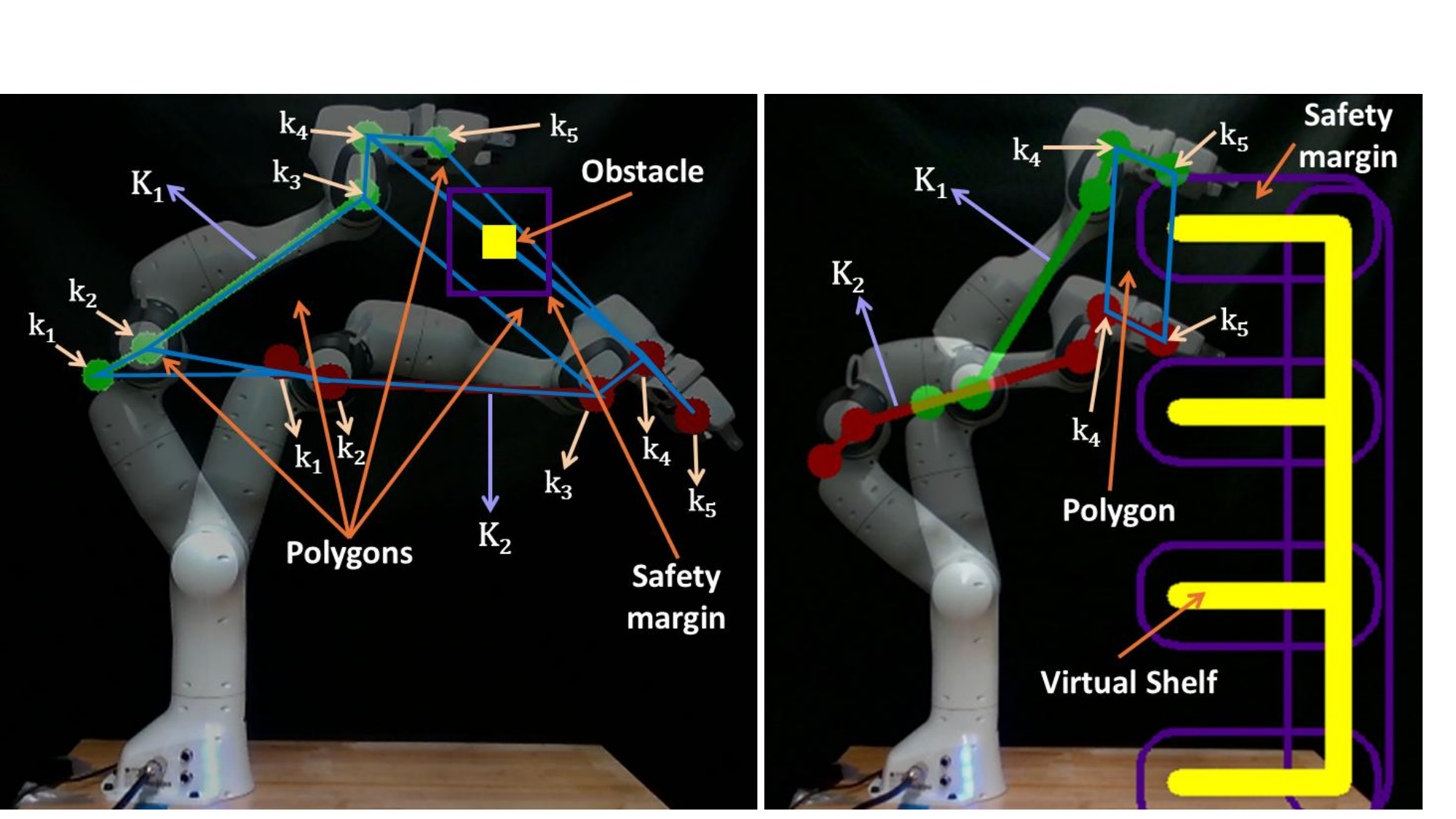}
      \caption{Polygon-based collision checking in image space. Four polygons (light blue) are defined by connecting consecutive keypoints—(\(\skey_1\), \(\skey_2\)), (\(\skey_2\), \(\skey_3\)), (\(\skey_3\), \(\skey_4\)), and (\(\skey_4\), \(\skey_5\))—from the current image state (\(\key_1\), green) to its neighbor (\(\key_2\), red). A collision is detected if any polygon intersects the obstacle contour (yellow) inflated by a purple safety margin, as seen for ([\(\skey_2\), \(\skey_3\)], [\(\skey_3\), \(\skey_4\)], and [\(\skey_4\), \(\skey_5\)]) in the first image and [\(\skey_4\), \(\skey_5\)] in the shelf scenario.}
      \vspace{-0.5em}
      \label{line-seg}
\end{figure}
\vspace{-1.75em}
\subsection{Adaptive Visual Servoing}
\vspace{-0.3em}
This work builds on the adaptive visual servoing method described in \cite{gandhi2022skeleton}, using a roadmap of collision-free sequence of goal image states. At each goal, vector of keypoints (\skey) in image state (\key) is tracked as visual features as described in \cite{chatterjee2023keypoints}. The controller moves the arm minimizing the feature error, computed as the difference between the current and the target \key. The Jacobian matrix, estimated online via least-square optimization of recent joint velocities and keypoints vector over a moving window, eliminates the need to read joint position from  encoder. This makes the control pipeline completely model-free. To improve accuracy, we reset the Jacobian estimation window at each new target keeping the estimate unbiased and relevant to the current goal. Since goal spacing in the image may vary across experiments, a saturation limit prevents velocity spikes, ensuring smooth motion and protecting the motor.

\begin{figure}[t]
      \centering
      \includegraphics[width=0.9\columnwidth]{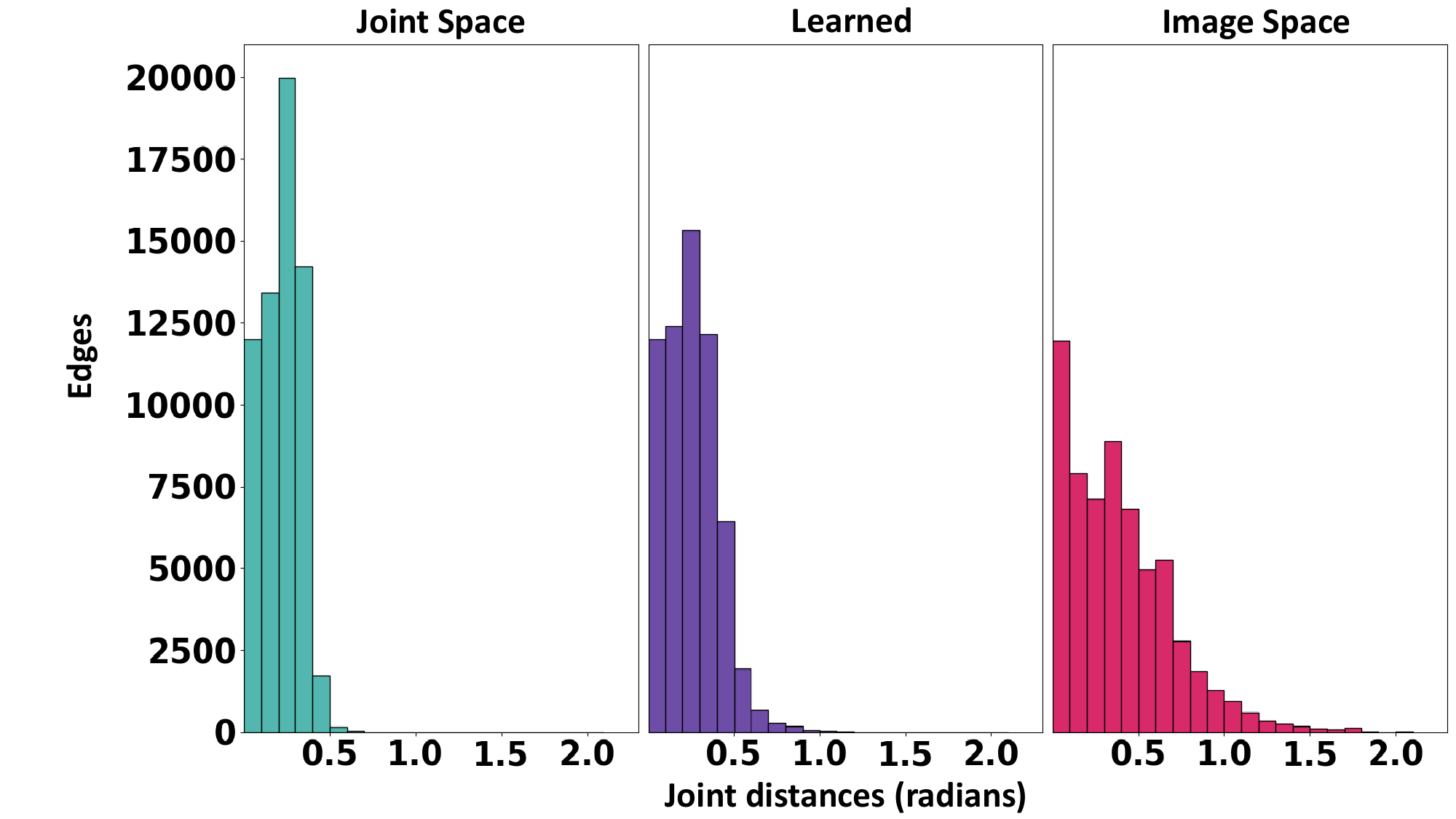}
      \caption{Distribution of joint displacements along roadmap edges. The \textbf{Learned} roadmap closely aligns with the \textbf{Joint Space} distribution, indicating accurate distance estimation. In contrast, the \textbf{Image Space} roadmap shows a wider spread, reflecting poor alignment between image and joint space.}
      \vspace{-1.5em}
      \label{histogram}
\end{figure}
\vspace{-1.0em}
\section{Experiments and Observations:}
\vspace{-0.45em}
We experimentally assessed the performance of the proposed vision only motion planning framework on a Franka Emika Panda Arm \cite{haddadin2024franka}.

\vspace{-1.25em}

\subsection{Experimental details}
\vspace{-0.3em}
\subsubsection{Generating Samples} We generated the required image state samples by collecting a finite dataset created by actuating the planar joints (Joints $2,4,6$) of the Franka arm, using velocities computed from \autoref{comp_vel} as explained in \autoref{ssec:dataset}, covering the robot's planar workspace.

\subsubsection{ Building the Roadmap} The collected samples \key were used as nodes to build roadmaps based on the distance metrics described in \autoref{ssec:dist_metric}. We refer to the roadmap using image space distances as the \textbf{Image Space} roadmap, and the one using learned distances as the \textbf{Learned} roadmap. For benchmarking, we also construct a \textbf{Joint Space} roadmap using actual joint displacements from encoders, used only for comparison, while preserving the model-free design. The k-neighbor value for all approaches was set to $25$. The first row of \autoref{rm_path_gen_time} shows the time required to generate each roadmap. The \textbf{learned} roadmap takes significantly longer to construct due to the additional computational overhead of running neural network inference during nearest neighbor computation.

\subsubsection{Obstacle Representation and Path Planning} In the experimental setup, obstacles were modeled as virtual shapes such as a rectangle or a triangle or a circle with a safety margin, as shown in the first image of \autoref{line-seg}. Paths were generated offline for various start (\keys) and goal (\keyg) incorporating the obstacle avoidance logic from \autoref{ssec:line-check}. These paths were later used in adaptive visual servoing experiments.

\subsubsection{Control Experiment Set-up for Single Obstacle} 
 The real-time control experiments used an Intel Realsense D435i camera in an eye-to-hand setup for visual feedback. The Panda arm followed the planned paths in $16$ obstacle-free and $10$ obstacle-avoidance experiments. The performance of each proposed roadmap was evaluated by comparing the joint position changes between intermediate image states to those of the \textbf{joint space} roadmap. The controller’s ability to guide the arm along collision-free paths was evaluated for efficiency and effectiveness.

\subsubsection{Control Experiment Set-up for Multiple Obstacles}\label{ssec:multi_obs}
To further evaluate the robustness of our planning method, we conducted $4$ additional experiments involving scenes with multiple obstacles of varying shapes and sizes. The \textbf{learned} roadmap successfully generated feasible paths in all $4$ scenarios. In contrast, the \textbf{image space} roadmap was able to find valid paths in only $2$ out of $4$ cases, failing in the remaining two due to limitations in its Euclidean distance metric when navigating tighter environments. 

\subsubsection{Control Experiment Set-up for Virtual Shelf}
\label{ssec:shelf}
We conducted a proof-of-concept experiment simulating a real-world object (a multi-level shelf), as shown in the second image of \autoref{line-seg}, and planned paths using both the \textbf{learned} and \textbf{image space} roadmaps.
\vspace{-1.25em}
\subsection{Roadmap and Path Planning Experiments}
\vspace{-0.35em}
In this section, we analyze the joint displacements along the edges of the three roadmaps to evaluate their efficiency. Path planning experiments were conducted in both collision-free and obstacle-avoidance scenarios to compare the joint distances covered by the paths generated from each roadmap.

\subsubsection{Distribution of Joint Distances of Edges for Different Roadmaps}
\autoref{histogram} shows the joint displacement histograms for the three roadmaps. The \textbf{learned} roadmap closely matches the \textbf{joint space} roadmap, with only minor deviations, indicating accurate estimation of joint displacements. In contrast, the \textbf{image space} roadmap shows a wider spread and larger displacements, suggesting less efficient transitions.

\subsubsection{Average Joint Distances for Planned Collision-Free Paths}
\label{random-rm}
We randomly selected $100$ \keys and \keyg pairs from the roadmaps for path planning without obstacles. As shown in \autoref{random_trials}, the \textbf{learned} roadmap produces joint distances closer to the \textbf{joint space} baseline, while the \textbf{image space} roadmap results in significantly higher displacements. This indicates that the \textbf{learned} roadmap generates more efficient paths, better suited for control convergence.

\begin{table}[t]
\caption{Roadmap building and average path generation time (s)}
\label{rm_path_gen_time}
\vspace{-0.5em}
\centering
\resizebox{0.475\textwidth}{!}{%
\begin{tabular}{|>{\centering\arraybackslash}p{3.5cm}|>{\centering\arraybackslash}p{2cm}|>{\centering\arraybackslash}p{2cm}|>{\centering\arraybackslash}p{2cm}|}
\hline
& \textbf{Joint Space} & \textbf{Learned} & \textbf{Image Space} \\
\hline
\hline
\textbf{Roadmap creation time (s)} & $\textbf{0.25}$ & $\textbf{766.19}$ & $\textbf{0.28}$ \\
\hline
% \textbf{Path} & \boldmath$2.422 \pm 2.46$ & \boldmath$3.19 \pm 2.57$ & \boldmath$2.38 \pm 2.4$ \\
\textbf{Query path time (s)} & \boldmath$2.422$ & \boldmath$3.19$ & \boldmath$2.38$ \\
\hline
\end{tabular}
}
\vspace{-0.5em}
\end{table}

\begin{table}[t]
\caption{Average joint distances (radians) for collision-free paths}
\label{random_trials}
\vspace{-0.5em}
\centering
\resizebox{0.475\textwidth}{!}{%
\begin{tabular}{|>{\centering\arraybackslash}p{3cm}|>{\centering\arraybackslash}p{2.5cm}||>{\centering\arraybackslash}p{2.5cm}|>{\centering\arraybackslash}p{2.5cm}|}
\hline
& \textbf{Joint Space} & \textbf{Learned} & \textbf{Image Space} \\
\hline
\hline
\textbf{Mean (radians)} & $\textbf{1.74}$ & $\textbf{2.19}$ & $\textbf{3.06}$ \\
\hline
\end{tabular}
}
\vspace{-1.75em}
\end{table}

\subsubsection{Planned paths for Control Experiments}

\label{ssec:path_planning}
We generated planned paths for two scenarios: $16$ start and goal pairs without checking for collision and $10$ pairs with collision avoidance. Paths were computed using the three roadmaps: \textbf{joint space}, \textbf{learned}, and \textbf{image space}. These precomputed paths were used in the control experiments described in \autoref{all_control_exps}. The second row of \autoref{rm_path_gen_time} reports the average time each roadmap took to generate paths for these scenarios.

For each planned path, we calculated metrics including the average number of waypoints (intermediate image states), joint distances between waypoints, total joint distances for the entire path, and Euclidean distances between image states (keypoint distances) both between waypoints and across the entire path. 

As observed in \autoref{exps_free_and_obs} the \textbf{learned} roadmap consistently resulted in fewer waypoints and shorter joint distances compared to the \textbf{image space} roadmap, which prioritizes minimizing keypoints distances in image space but incurs higher joint displacements.

Notably, the joint distances required to traverse $1000$ pixels in image space were much higher for the \textbf{image space} roadmap than for the \textbf{learned} roadmap. This suggests that reliance on \textbf{image space} proximity may lead to less efficient joint-space paths.

\begin{table}[t]
\caption{Comparison of joint distances and keypoints distances in image space over experiments}
\label{exps_free_and_obs}
\centering
\resizebox{0.45\textwidth}{!}{%
\begin{tabular}{|
>{\centering\arraybackslash}m{2cm}||
>{\centering\arraybackslash}m{1cm}|
>{\centering\arraybackslash}m{1cm}|
>{\centering\arraybackslash}m{1cm}||
>{\centering\arraybackslash}m{1cm}|
>{\centering\arraybackslash}m{1cm}|
>{\centering\arraybackslash}m{1cm}|}
\hline
& \multicolumn{3}{c||}{\textbf{Without Obstacle Avoidance}} & \multicolumn{3}{c|}{\textbf{With Obstacle Avoidance}} \\
\hline
\textbf{Roadmaps} & \textbf{Joint Space} & \textbf{Learned} & \textbf{Image Space} & \textbf{Joint Space} & \textbf{Learned} & \textbf{Image Space} \\
\hline
\makecell{\textbf{Number of} \\ \textbf{Experiments}} & \textbf{16} & \textbf{16} & \textbf{16} & \textbf{10} & \textbf{10} & \textbf{10} \\
\hline
\makecell{\textbf{Avg. No. of} \\ \textbf{Waypoints}} & \textbf{8} & \textbf{8} & \textbf{14} & \textbf{11} & \textbf{13} & \textbf{15} \\
\hline
\textbf{Avg. Joint Distances (radians) b/w Waypoints} & \textbf{0.25} & \textbf{0.3} & \textbf{0.32} & \textbf{0.28} & \textbf{0.28} & \textbf{0.35} \\
\hline
\textbf{Avg. Keypoints Distances (pixels) b/w Waypoints} & \textbf{167.99} & \textbf{174.76} & \textbf{84.11} & \textbf{139.53} & \textbf{129.13} & \textbf{89.36} \\
\hline
\textbf{Avg. Joint Distances (radians) over Entire Path} & \textbf{1.92} & \textbf{2.27} & \textbf{4.29} & \textbf{3.04} & \textbf{3.46} & \textbf{5.36} \\
\hline
\textbf{Avg. Keypoints Distances (pixels) over Entire Path} & \textbf{1222.42} & \textbf{1278.92} & \textbf{1157.00} & \textbf{1532.72} & \textbf{1569.37} & \textbf{1361.43} \\
\hline
\textbf{Joint Distance (radians) Traversed To Move 1000 pixels in image space} & \textbf{1.6} & \textbf{1.77} & \textbf{3.79} & \textbf{1.98} & \textbf{2.19} & \textbf{3.9} \\
\hline
\end{tabular}
} % End of resizebox
\vspace{-1.75em}
\end{table}

We have two theories from the above observations:

\begin{itemize}
\item The \textbf{image space} roadmap uses Euclidean distances, leading A* to favor shorter pixel paths with more intermediate waypoints. In contrast, the \textbf{learned} roadmap, based on joint displacements, yields more direct paths with fewer waypoints.
\item Image states that are close in pixel space can be far apart in \textbf{joint space}, as shown in \autoref{exps_free_and_obs}. This can result in larger joint movements and control errors, especially in systems with non-holonomic constraints.
\end{itemize}

\vspace{-1.25em}
\subsection{Control Experiments}
\vspace{-0.35em}
\label{all_control_exps}
The adaptive visual servoing experiments\footnote{Planning and control experiments including apf videos, are available at \href{https://drive.google.com/file/d/1rnBGJeI8it6Vs-4VtHuLjEVLhfE7i4lc}{this link}. The details of how to use the link is in the supplementary ReadMe file. The supplementary \href{https://www.youtube.com/watch?v=T-9qihoV6JQ}{video} and relevant \href{https://github.com/JaniC-WPI/KPDataGenerator}{codes} can be found in respective links} used precomputed paths from  \autoref{ssec:path_planning}, with control gains optimized to minimize rise and settling times while keeping overshoot within 5\%, by careful tuning. We also performed adaptive visual-servoing experiments for the $4$ scenarios for \textbf{learned} and \textbf{image space} roadmaps as defined in \autoref{ssec:multi_obs}. To highlight the advantages of planning-based control over reactive methods, we implemented an Artificial Potential Field (APF) controller \cite{choset2005principles} and tested it on the same $4$ multi-obstacle scenarios from \autoref{ssec:multi_obs}.

The control metric in \autoref{performance_data} shows that the \textbf{image space} roadmap succeeded in $69.2$\% of cases, while the \textbf{learned} roadmap achieved $100$\% success. When successful, the \textbf{image space} roadmap had faster transients. This can be attributed to its shorter pixel-distance paths, which reduce image space error more quickly in scenarios where joint displacements are relatively smaller. \autoref{exps_success_failure} uses the same metrics as \autoref{exps_free_and_obs}, grouped by success and failure.

\textbf{Image space} roadmaps fail when the robot tries to follow short pixel paths that require large joint movements. Due to non-holonomic constraints, the robot might need to traverse a curved path in \textbf{joint space} to achieve even a small shift in \textbf{image space}, risking collisions or workspace violations. \textbf{Learned} roadmaps avoid this by producing motions that better match the robot’s capabilities.

 \autoref{performance_data_multi_obs} depicts the performance of \textbf{learned} and \textbf{image space} roadmaps in multi-obstacle scenarios. While \textbf{image space} failed to plan in $50$\% of the cases, all successful plans led to successful control; the \textbf{learned} roadmap succeeded in both planning and control for all cases. The APF controller failed to converge in all scenarios.

 Despite its simplicity, APF failed in all cases, where the robot started between closely spaced obstacles, getting trapped in local minima and oscillating between attractive and opposing repulsive forces, without reaching the goal. These failures highlight the limitations of reactive methods in cluttered scenes and the necessity of global planners like our roadmap-based approach for collision avoidance.

\autoref{performance_data_shelf} presents control performance in a proof-of-concept scenario involving simulated real-world object, as described in \autoref{ssec:shelf}. Here, the path planned by the \textbf{image space} roadmap led to a collision with the object, despite a defined safety margin. Further experiments are needed to conclusively determine why the robot collided despite a collision-free planned path.

\begin{table}[t]
\caption{Performance results for the control experiments with and without obstacle avoidance}
\label{performance_data}
\centering
\resizebox{0.475\textwidth}{!}{%
\begin{tabular}{
|>{\centering\arraybackslash}m{2cm}||
 >{\centering\arraybackslash}m{1cm}|
 >{\centering\arraybackslash}m{1cm}|
 >{\centering\arraybackslash}m{1cm}||
 >{\centering\arraybackslash}m{1cm}|
 >{\centering\arraybackslash}m{1cm}|
 >{\centering\arraybackslash}m{1cm}|
}
\hline
& \multicolumn{3}{c||}{\textbf{Without Obstacle Avoidance}} & \multicolumn{3}{c|}{\textbf{With Obstcale Avoidance}} \\
\hline
\textbf{Performance Metrics} & \textbf{Joint Space} & \textbf{Learned} & \textbf{Image Space} & \textbf{Joint Space} & \textbf{Learned} & \textbf{Image Space} \\
\hline
\hline
\textbf{Successful Experiments} & \textbf{16/16} & \textbf{16/16} & \textbf{13/16} & \textbf{10/10} & \textbf{10/10} & \textbf{5/10} \\
\hline
\textbf{System Rise time (s)} & \textbf{74.92} & \textbf{97.53} & \textbf{94.75} & \textbf{105.38} & \textbf{125.38} & \textbf{90.46} \\
\hline
\textbf{System Settling time (s)} & \textbf{94.37} & \textbf{118.62} & \textbf{101.99} & \textbf{118.18} & \textbf{155.14} & \textbf{96.44} \\
\hline
\textbf{End effector Rise time (s)} & \textbf{74.86} & \textbf{97.53} & \textbf{94.39} & \textbf{105.02} & \textbf{125.38} & \textbf{90.46} \\
\hline
\textbf{End effector Settling time (s)} & \textbf{94.37} & \textbf{114.02} & \textbf{99.81} & \textbf{118.18} & \textbf{147.14} & \textbf{94.22} \\
\hline
\textbf{Overshoot (\%)} & \textbf{1.94} & \textbf{2.61} & \textbf{1.89} & \textbf{1.93} & \textbf{2.36} & \textbf{1.35} \\
\hline
\textbf{Execution time (s)} & \textbf{125.17} & \textbf{148.92} & \textbf{140.78} & \textbf{146.02} & \textbf{201.68} & \textbf{124.70} \\
\hline
\end{tabular}
}
\end{table}

\begin{table}[t]
\caption{Comparison of joint distances and keypoints distances in image space over experiments for successful and failed experiments}
\label{exps_success_failure}
\centering
\resizebox{0.45\textwidth}{!}{%
\begin{tabular}{|
>{\centering\arraybackslash}m{2cm}||
>{\centering\arraybackslash}m{1cm}|
>{\centering\arraybackslash}m{1cm}|
>{\centering\arraybackslash}m{1cm}||
>{\centering\arraybackslash}m{1cm}|
>{\centering\arraybackslash}m{1cm}|
>{\centering\arraybackslash}m{1cm}|}
\hline
& \multicolumn{3}{c||}{\textbf{Successful}} & \multicolumn{3}{c|}{\textbf{Failed (Image Space)}} \\
\hline
\textbf{Roadmaps} & \textbf{Joint Space} & \textbf{Learned} & \textbf{Image Space} & \textbf{Joint Space} & \textbf{Learned} & \textbf{Image Space} \\
\hline
\textbf{Number of Experiments} & \textbf{18} & \textbf{18} & \textbf{18} & \textbf{8} & \textbf{8} & \textbf{8} \\
\hline
\textbf{Avg No. of Waypoints} & \textbf{8} & \textbf{8} & \textbf{14} & \textbf{11} & \textbf{13} & \textbf{15} \\
\hline
\textbf{Avg. Joint Distances (radians) b/w Waypoints} & \textbf{0.25} & \textbf{0.3} & \textbf{0.32} & \textbf{0.27} & \textbf{0.29} & \textbf{0.35} \\
\hline
\textbf{Avg. Keypoints Distances (pixels) b/w Waypoints} & \textbf{166.77} & \textbf{163.41} & \textbf{86.03} & \textbf{135.17} & \textbf{143.24} & \textbf{86.35} \\
\hline
\textbf{Avg. Joint Distances (radians) over Entire Path} & \textbf{2.10} & \textbf{2.37} & \textbf{4.42} & \textbf{2.92} & \textbf{3.52} & \textbf{5.33} \\
\hline
\textbf{Avg. Keypoints Distances (pixels) over Entire Path} & \textbf{1307.02} & \textbf{1257.02} & \textbf{1210.93} & \textbf{1419.93} & \textbf{1691.26} & \textbf{1291.2} \\
\hline
\textbf{Joint Distance (radians) Traversed To Move 1000 pixels in image space} & \textbf{1.6} & \textbf{1.9} & \textbf{3.71} & \textbf{2.1} & \textbf{2.1} & \textbf{4.12} \\
\hline
\end{tabular}
}
\vspace{-1.05em}
\end{table}

\begin{table}[t]
\caption{Performance comparison of learned and image space roadmaps and APF with multiple obstacles}
\label{performance_data_multi_obs}
\centering
\renewcommand{\arraystretch}{1.3}  % Slightly increase row height
\resizebox{0.5\textwidth}{!}{%     % Slightly narrower table
\begin{tabular}{|>{\centering\arraybackslash}m{2cm}||>{\centering\arraybackslash}m{1.2cm}|>{\centering\arraybackslash}m{1.2cm}|>{\centering\arraybackslash}m{1.2cm}|>{\centering\arraybackslash}m{1.2cm}|>{\centering\arraybackslash}m{1.2cm}|>{\centering\arraybackslash}m{1.2cm}|>{\centering\arraybackslash}m{1.1cm}|>{\centering\arraybackslash}m{1.0cm}|}
\hline
\textbf{Roadmap} & \textbf{Successful Planning.} & \textbf{Successful Control.} & \textbf{System Rise Time (s)} & \textbf{System Settling Time (s)} & \textbf{EE Rise Time (s)} & \textbf{EE Settling Time (s)} & \textbf{Overshoot (\%)} & \textbf{Exec. Time (s)} \\
\hline \hline
\textbf{Learned} & \textbf{4/4} & \textbf{4/4} & \textbf{96.53} & \textbf{99.68} & \textbf{95.98} & \textbf{97.83} & \textbf{0.87} & \textbf{121.65} \\
\hline
\textbf{Image Space} & \textbf{2/4} & \textbf{2/2} & \textbf{89.1} & \textbf{102.6} & \textbf{89.1} & \textbf{100.05} & \textbf{0.43} & \textbf{122.45} \\
\hline
\textbf{APF} & \textbf{NA} & \textbf{0} & \textbf{NA} & \textbf{NA} & \textbf{NA} & \textbf{NA} & \textbf{NA} & \textbf{NA} \\
\hline
\end{tabular}
}
\vspace{-0.75em}
\end{table}

\begin{table}[t]
\caption{Performance comparison for learned and image space roadmaps with proof-of-concept real-object scenario}
\label{performance_data_shelf}
\centering
\renewcommand{\arraystretch}{1.3}  % Slightly increase row height
\resizebox{0.5\textwidth}{!}{%     % Slightly narrower table
\begin{tabular}{|>{\centering\arraybackslash}m{2cm}||>{\centering\arraybackslash}m{1.2cm}|>{\centering\arraybackslash}m{1.2cm}|>{\centering\arraybackslash}m{1.2cm}|>{\centering\arraybackslash}m{1.2cm}|>{\centering\arraybackslash}m{1.2cm}|>{\centering\arraybackslash}m{1.2cm}|>{\centering\arraybackslash}m{1.1cm}|>{\centering\arraybackslash}m{1.0cm}|}
\hline
\textbf{Roadmap} & \textbf{Successful Planning.} & \textbf{Successful Control.} & \textbf{System Rise Time (s)} & \textbf{System Settling Time (s)} & \textbf{EE Rise Time (s)} & \textbf{EE Settling Time (s)} & \textbf{Overshoot (\%)} & \textbf{Exec. Time (s)} \\
\hline \hline
\textbf{Learned} & \textbf{1/1} & \textbf{1/1} & \textbf{34.8} & \textbf{75.8} & \textbf{28.9} & \textbf{75.8} & \textbf{3.24} & \textbf{97.5} \\
\hline
\textbf{Image Space} & \textbf{1/1} & \textbf{0/1} & \textbf{NA} & \textbf{NA} & \textbf{NA} & \textbf{NA} & \textbf{NA} & \textbf{NA} \\
\hline
\end{tabular}
}
\vspace{-1.0em}
\end{table}

In summary, both roadmaps exhibit unique benefits. The \textbf{image space} roadmap enables faster execution when successful but lacks reliability, while the \textbf{learned} roadmap offers greater robustness by leveraging joint displacement-like distances. The choice depends on the application context.
\vspace*{-1.0em}
\section{Conclusion and Future Work}
\vspace{-0.35em}
In conclusion, this work introduced a novel framework for collision-free motion planning of robotic manipulators that relied solely on visual features, eliminating the need for explicit robot models or encoder feedback. 

The \textbf{learned} roadmap offered smoother, more reliable transitions, and due to its joint displacement-based distance definition, the paths it generated maintained joint-space holonomy when it existed. In contrast, the paths produced by the \textbf{image space} roadmap sometimes failed to maintain joint-space holonomy, even when holonomy existed in the \textbf{image space}. However, the \textbf{image space} roadmap provided faster transient responses and simplicity, making it advantageous for applications where speed and computational efficiency were prioritized.

As a future direction, to extend our approach to higher DoF and out-of-plane motion, we plan to reuse the same data collection pipeline, existing data collection pipeline while capturing trajectories that span the full 3D workspace. To ensure safe interaction with real-world obstacles, we plan to estimate monocular depth replacing the 2D safety polygon with a 3D bounding box. To improve keypoint robustness under environmental and self-occlusion, we aim to integrate a graph-based model to enforce spatial consistency within the keypoint detection network. We are also exploring the extension of this pipeline to soft and continuum robotic systems, specifically a soft origami arm. Additionally, We are also investigating a quantitative formulation to characterize and analyze the role of non-holonomic constraints in image-based motion planning.

\vspace*{-1.0em}

\bibliographystyle{IEEEtran}
% \bibliography{IEEEabrv,bibliography}

\end{document}